\documentclass[10pt,twocolumn,letterpaper]{article}

\usepackage{sty}
\usepackage{times}
\usepackage{epsfig}
\usepackage{graphicx}
\usepackage{amsmath}
\usepackage{amssymb}
\usepackage{multirow}
\usepackage{booktabs} 
\usepackage{tabularx}
\usepackage{caption}
\usepackage{subcaption}

\usepackage{pifont}
\newcommand{\cmark}{\ding{51}}%
\newcommand{\xmark}{\ding{55}}

\DeclareMathOperator*{\argmax}{arg\,max}

\newcommand{\ptr}{\mathrm{P}_{\mathrm{tr}}}
\newcommand{\pte}{\mathrm{P}_{\mathrm{te}}}

\newcommand{\Lagr}{\mathcal{L}}
\newcommand{\score}{\mathrm{s}}
\newcommand{\knownd}{d^*}
\newcolumntype{?}[0]{!{\vrule width 0.05mm}}

\usepackage{dsfont}

\newcommand\blfootnote[1]{%
  \begingroup
  \renewcommand\thefootnote{}\footnote{#1}%
  \addtocounter{footnote}{-1}%
  \endgroup
}

\usepackage[breaklinks=true,bookmarks=false]{hyperref}

\vcvcfinalcopy 

\begin{document}

\title{Fair Visual Recognition in Limited Data Regime\\ using Self-Supervision and Self-Distillation}

\author{Pratik Mazumder$^{1,\ast}$, \hspace{0.5cm} Pravendra Singh$^{2,\ast}$, \hspace{0.5cm} Vinay P. Namboodiri$^{1,3}$ \\
$^1$IIT Kanpur, India\hspace{0.5cm}
$^2$IIT Roorkee, India\hspace{0.5cm}
$^3$University of Bath, United Kingdom\\
{\tt\small pratikm@cse.iitk.ac.in, pravendra.singh@cs.iitr.ac.in, vpn22@bath.ac.uk}
}

\maketitle

\ifvcvcfinal\thispagestyle{empty}\fi

\begin{abstract} 
Deep learning models generally learn the biases present in the training data. Researchers have proposed several approaches to mitigate such biases and make the model fair. Bias mitigation techniques assume that a sufficiently large number of training examples are present. However, we observe that if the training data is limited, then the effectiveness of bias mitigation methods is severely degraded. In this paper, we propose a novel approach to address this problem. Specifically, we adapt self-supervision and self-distillation to reduce the impact of biases on the model in this setting. Self-supervision and self-distillation are not used for bias mitigation. However, through this work, we demonstrate for the first time that these techniques are very effective in bias mitigation. We empirically show that our approach can significantly reduce the biases learned by the model. Further, we experimentally demonstrate that our approach is complementary to other bias mitigation strategies. Our approach significantly improves their performance and further reduces the model biases in the limited data regime. Specifically, on the L-CIFAR-10S skewed dataset, our approach significantly reduces the bias score of the baseline model by 78.22\% and outperforms it in terms of accuracy by a significant absolute margin of 8.89\%. It also significantly reduces the bias score for the state-of-the-art domain independent bias mitigation method by 59.26\% and improves its performance by a significant absolute margin of 7.08\%.
\end{abstract}

\section{Introduction} 
\blfootnote{$^\ast$ The first two authors contributed equally.}
Deep learning models have been successfully used to solve several real-world problems, but they need to be trained on a large amount of labeled data. The training data often contains biases related to color, gender, appearance, and others. Standard deep learning training techniques capture such biases, and the resulting models make biased predictions. Researchers have proposed several approaches to mitigate bias in the deep learning models. These approaches assume the availability of large amounts of training data. In this paper, we empirically demonstrate how a limited training data severely increases the impact of biases on the model predictions. We also demonstrate that the effectiveness of bias mitigation approaches is significantly reduced if the training data is limited. We propose a novel approach for addressing this problem that can also complement other bias mitigation strategies and improve their performance significantly.

The biases in model predictions are generally due to the same biases in the training data. This problem arises when the model finds an unwanted correlation in the data that minimizes the training loss. For example, if in a dataset the majority of the nurses are female, the model may become more inclined towards predicting a woman as a nurse compared to a man since the loss incurred by the model will be minimized. Further, if the dataset contains limited training data, any such biases will get reinforced. Therefore, the model may produce even more biased predictions when trained on limited data. 

Some of the popular approaches for bias mitigation are the domain adversarial training~\cite{tzeng2015simultaneous,ryu2017improving,alvi2018turning}, Reducing Bias Amplification~\cite{zhao_men_2017}, domain discriminative training \cite{wang2020towards} and domain independent training \cite{wang2020towards} approaches. The authors in \cite{wang2020towards} propose benchmark datasets for bias mitigation by deliberately introducing biases in the form of color/grayscale images, out-of-distribution images, scaled images, and cropped images in some classes. We propose limited data versions of these benchmark datasets~\cite{wang2020towards} and study the performance of bias mitigation methods in the limited data regime.

 We propose a novel approach that uses self-supervision and self-distillation to mitigate the bias in the model. Our approach involves training the network on an auxiliary self-supervision task and applying the self-distillation technique to the model. Self-supervision and self-distillation are generally used for improving the representation learning, and generalization ability of the model and are not regarded as bias mitigation approaches. The \textit{novelty} of this work lies in adapting these two techniques for this setting and demonstrating that these techniques are very effective in bias mitigation (see Sec.~\ref{sec:motivation}). We empirically show that our approach significantly reduces the impact of biases in the network. Specifically, on the L-CIFAR-10S dataset, our approach significantly reduces the bias score of the baseline model by 78.22\%. To the best of our knowledge, this is the \textit{first work} to use such an approach to improve fairness. Further, our approach can be seamlessly integrated with other bias mitigation approaches and can significantly boost their bias mitigation ability in this setting. For example, on the L-CIFAR-10S dataset, our approach significantly reduces the bias score for the strategic sampling, adversarial, domain discriminative, and domain independent bias mitigation approaches by 89.81\%, 82.04\%, 65.95\%, and 59.26\%, respectively. We also demonstrate in Sec.~\ref{sec:performsufdata} that our approach improves the model performance even when sufficient training data is present. We perform extensive ablations to validate our approach.
 
Our contributions can be summarized as follows:
\begin{itemize}
    \item We empirically show that the effect of biases becomes even more adverse when the training data is limited.
    
    \item We empirically demonstrate that the performance of bias mitigation approaches suffers significantly when the available training data is limited.
    
    \item We demonstrate for the first time that self-distillation and self-supervision are very effective in bias mitigation. We propose a novel approach that adapts these techniques for mitigating bias in the limited data setting. 
    
    \item  We perform experiments on several benchmark datasets to validate our approach. We empirically show that our approach can augment other bias mitigation methods and significantly improve their performance in the limited data setting.
\end{itemize}

\section{Related Works}

\subsection{Bias Mitigation Approaches}
Researchers have proposed various approaches for dealing with biases arising from erroneous correlations present in the training dataset~\cite{anne2018women, grover2019bias, wang2019balanced, kim2019learning, li2019repair, quadrianto2019discovering, wang2019racial, grover2019fair}. The classes/attributes that are responsible for the spurious correlations are referred to as protected attributes. The methods proposed in \cite{khosla2012undoing, zemel2013learning} aim to mitigate bias in simple linear models. The authors in \cite{zhao_men_2017} propose to mitigate bias by updating the inference to match a target distribution. InclusiveFaceNet~\cite{ryu2018inclusivefacenet} utilizes demographic information prior to perform better and less biased attribute detection. The domain discriminative approach described in \cite{wang2020towards} is also based on this approach and creates a new set of classes that incorporates the protected attribute/class. The method in \cite{dwork2018decoupled} decouples classifiers in order to reduce the effect of biases on the network predictions. The authors in \cite{wang2020towards} propose the domain independent bias mitigation strategy that uses the same principle. This approach generates separate sets of predictions for each value of the protected domain attribute to reduce the bias and then combines these predictions. Several researchers have also proposed an adversarial approach of bias mitigation \cite{wang2020towards,alvi2018turning,zhang2018mitigating,edwards2016censoring,ganin2015unsupervised} that trains the network to reduce its ability to identify the protected attribute/class. 

The authors in \cite{wang2020towards} propose benchmark datasets for comparing bias mitigation strategies by deliberately introducing biases in the form of color/grayscale images, out-of-distribution images, scaled images, and cropped images in some classes. The work in \cite{wang2020towards} also compares various bias mitigation approaches on these benchmark datasets.

However, we demonstrate that the biases in model predictions become worse with limited training data. We also empirically show that the effectiveness of bias mitigation approaches significantly diminishes in the case of limited training data.

\subsection{Self-Supervision}
Researchers have proposed several self-supervision approaches to improve representation learning of the network without requiring additional labeled data. Self-supervision methods use techniques like modifying images to create artificial labels, modeling invariance through contrastive learning, and others \cite{dosovitskiy2014discriminative,zhang2016colorful}. The authors in \cite{gidaris2018unsupervised} propose to rotate the image and train the network to predict the rotation angle as the self-supervised task. 

Contrastive learning based self-supervised tasks train the network to bring similar images closer and push dissimilar images away in the feature space. SimCLR \cite{chen2020simple} applies random data augmentations to produce two different views of the same image and performs contrastive learning using these views, and also uses the negative samples present in the batch. The authors in \cite{chen2020exploring} propose the state-of-the-art SimSiam self-supervision technique. SimSiam also performs contrastive learning on two randomly augmented views of the same image without requiring negative samples or a large batch size. We use SimSiam as an auxiliary task in our approach to mitigate the bias in the network (see Sec.~\ref{sec:motivation}). We empirically show in Sec.~\ref{sec:ablchoicess} that SimSiam outperforms other self-supervised tasks when used as an auxiliary task in our approach.

\subsection{Self-Distillation}
The authors in \cite{hinton2015distill} propose to use knowledge distillation for knowledge transfer from a teacher network to a student network. This process involves training the student network to match the output logits/soft predictions of the teacher network in addition to the training objective of the primary task. When the teacher and student architectures are the same, the knowledge distillation process is referred to as self-distillation. The authors in \cite{mobahi2020selfdistillation} demonstrate that self-distillation improves the test set performance of the network. The distillation process increases the generalization ability of the network without requiring additional labeled data for training. We use self-distillation in our approach to reduce the impact of biases in the model (see Sec.~\ref{sec:motivation}).

\section{Effect of Biased Data on Model Predictions}\label{sec:fullstory} 

The authors in \cite{wang2020towards} propose the CIFAR-10 Skewed (CIFAR-10S) dataset~\cite{wang2020towards} for comparing various bias mitigation strategies. CIFAR-10S retains the 10 classes and 50,000 training images of the CIFAR-10~\cite{krizhevsky2009learning} dataset but modifies the training images to introduce a bias using two domains, i.e., color and grayscale. The color domain has five classes, each containing 95\% color and 5\% grayscale images (95-5\% skew). The remaining five classes belong to the grayscale domain and contain 95\% grayscale and 5\% color images each (95-5\% skew). Therefore, the training data of the color and grayscale domain classes are erroneously skewed towards the colored images and grayscale images, respectively. As a result, the domain can be considered as a protected attribute, and the model performance has to be de-biased with respect to this attribute. For evaluating the model, CIFAR-10S has two separate test datasets that basically are the color and grayscale copies of the entire CIFAR-10 test set, respectively. The CIFAR-10S dataset is designed in such a way that the test accuracy is indicative of the model bias, which helps to easily compare the fairness of different bias mitigation approaches~\cite{wang2020towards}.

\subsection{L-CIFAR-10S Dataset}\label{sec:limiteddata}
We propose a limited data version (L-CIFAR-10S) of the CIFAR-10S dataset, obtained by reducing the number of images per class to 5\%. However, we want to preserve the skewness (95-5\%) in the dataset. Therefore, we choose the first 5\% images from each class while maintaining the same level of skew, i.e., we separately choose the first 5\% images from the color and grayscale images in that class. As a result, if a class contained 95\% color images and 5\% grayscale images or vice-versa, the ratio of color and grayscale images will still remain the same in L-CIFAR-10S. For evaluating the model, we use the same color and grayscale test sets of CIFAR-10S. Therefore, this new L-CIFAR-10S dataset is similar to the full-sized CIFAR-10S dataset except for the number of training images. As a result, similar to CIFAR-10S \cite{wang2020towards}, the test accuracy over L-CIFAR-10S is also indicative of the model bias.

\subsection{Effect of Biases with Limited vs. Full Data }\label{sec:story}
\begin{figure}
     \centering
     \begin{subfigure}[b]{0.23\textwidth}
         \centering
         \includegraphics[width=\textwidth]{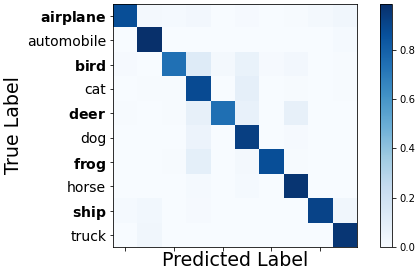}
         \caption{}
     \end{subfigure}
     \begin{subfigure}[b]{0.23\textwidth}
         \centering
         \includegraphics[width=\textwidth]{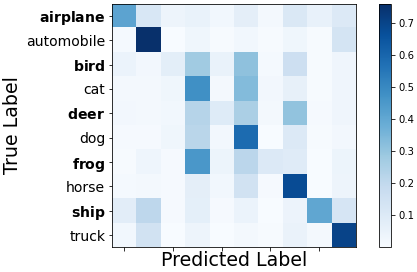}
         \caption{}
     \end{subfigure}
     
        \caption{Figure shows the confusion matrix of the colored test set predictions of a ResNet-18 model trained on (a) CIFAR-10S (b) L-CIFAR-10S. The confusion matrix indicates that when the training data is limited (L-CIFAR-10S) the biases present in the dataset have a significantly more severe effect on the predictions of the network. The bold labels denote the grayscale domain classes.}
        \label{fig:confusion}
        \vspace{-16pt}
\end{figure}

When we train the baseline ResNet-18 network on the L-CIFAR-10S dataset, the resulting model achieves 41.04\% (see Table \ref{table:cifar}) classification accuracy on the color test set. This is significantly lower as compared to the 89.0\% achieved by training on the CIFAR-10S dataset \cite{wang2020towards}. If we only consider this result, we may be tricked into assuming that this adverse effect on the baseline model accuracy is only due to the decrease in training data. However, if we convert all the training images in L-CIFAR-10S to grayscale and train the same ResNet-18 model using this data, the resulting model achieves a classification accuracy of 65.57\% on the colored test set (GTCT:GrayTrainColorTest). Similarly, ResNet-18 model trained on an all color version (details are given in the appendix) of L-CIFAR-10S achieves 66.53\% on the colored test set (CTCT:ColorTrainColorTest). Therefore, the bias in L-CIFAR-10S plays a major role in the significantly lower performance of the baseline (65.57/66.53\%$\rightarrow$41.04\%). This is also evident from the bias score (lower better) of the baseline, which increases from 0.074 (for full training data, i.e., CIFAR-10S) to 0.349 in the limited data setting, which is a massive increase by 371.62\%. Similar to the CIFAR-10S dataset, in the L-CIFAR-10S dataset, color domain classes have a majority of colored training images leading to an erroneous correlation in the model. But since the number of images in each class is low, the effect of this erroneous correlation is further magnified in the trained model, and it is even more inclined towards predicting one of the five color domain classes for any colored test image. As pointed out in \cite{wang2020towards}, ideally, the network should perform at least as well as GTCT and approach the performance of CTCT. The gap in the performance of the baseline model w.r.t. GTCT for the L-CIFAR-10S dataset due to the bias in the data is 24.53\%, which is significantly greater than the gap of 4\%~\cite{wang2020towards} w.r.t. GTCT (93.0\%) for the CIFAR-10S dataset (details in the appendix). Therefore, with limited training data, the impact of the bias on the model predictions has significantly worsened (as can also be seen in Fig.~\ref{fig:confusion}). Our approach trained on L-CIFAR-10S achieves a classification accuracy of $53.31\%$ (see Table \ref{table:cifar}) on the color test, which is significantly higher than the baseline and reduces this gap from 24.53\% to 12.26\%. Our approach also reduces the bias score of the baseline by 78.22\% (from 0.349 to 0.076). This indicates that the self-supervision and self-distillation used in our approach are very effective in bias mitigation.

\subsection{Effect of Biases on Bias Mitigation Approaches with Limited vs. Full Data }

We train bias mitigation approaches on the L-CIFAR-10S dataset. The domain discriminative training approach \cite{wang2020towards} achieves a classification accuracy of 41.47\% (see Table \ref{table:cifar}) on the colored test set of L-CIFAR-10S. Therefore, the gap in its performance compared to the GTCT for the L-CIFAR-10S dataset is 24.1\%, which is significantly greater than the gap of 1.8\%~\cite{wang2020towards}  w.r.t. GTCT (93.0\%) for the CIFAR-10S dataset (details in the appendix). In fact, the improvement due to the domain discriminative approach is negligible in this setting. Another important observation is that the bias score (lower better) of this approach increases from 0.040 (for full training data, i.e., CIFAR-10S) to 0.232 in the limited data setting, which is a huge increase by 480\%. The domain independent training approach \cite{wang2020towards} achieves a classification accuracy of 59.25\% (see Table \ref{table:cifar}) on the colored test set of L-CIFAR-10S. Therefore, the gap in its performance compared to GTCT for the L-CIFAR-10S dataset is 6.32\%, which is significantly greater than the gap of 0.6\%~\cite{wang2020towards} w.r.t. GTCT (93.0\%) for the CIFAR-10S dataset (details in the appendix). The domain independent approach also suffers from a 575\% increase in bias score in the limited data setting. Therefore, with limited training data, bias mitigation approaches become significantly less effective. The severe impact of limited training data on the effectiveness of bias mitigation approaches is a \textit{novel finding} of this work. The domain independent method augmented with our approach trained on L-CIFAR-10S achieves a classification accuracy of $66.42\%$ (see Table \ref{table:cifar}) on the color test set, which outperforms GTCT and is very close to CTCT for the L-CIFAR-10S dataset. Since our approach significantly reduces the bias score of the domain independent method by 59.26\% and improves its performance by a significant absolute margin of 7.17\%, the effectiveness of our approach in performing bias mitigation is validated.

\section{Proposed Approach} 
 
 \subsection{Motivation}\label{sec:motivation}
Our proposed approach, Limited Data Bias Mitigation approach (LDBM), uses self-supervision and self-distillation for bias mitigation. However, self-supervision and self-distillation are used for improving the generalization ability and representation quality of the model. Therefore, these techniques help the model learn useful/meaningful and discriminative features from the training data, thereby minimizing the impact of irrelevant features on the model. Consequently, if the training data suffers from biases due to any unwanted correlations, these techniques will also prevent the model from learning such biases, which is the objective of bias mitigation. In the limited data setting, this problem becomes even more severe, and any biases in the model, including those resulting from unwanted correlations in the training data, will get reinforced. This is why the baseline model and other bias mitigation approaches suffer from a significant impact of biases in this setting. However, due to the reasons mentioned above, our approach is effective in mitigating various types of biases in the baseline model and also significantly improves the effectiveness of the bias mitigation approaches in this setting. The self-supervised auxiliary task and the self-distillation process used in our approach are described as follows.

\subsection{Training with a Self-Supervised Auxiliary Task}
The authors in \cite{chen2020exploring} explain that SimSiam uses a siamese architecture for modeling invariance, which is a focus of representation learning. We employ a self-supervision training technique as an auxiliary task in our approach in order to reduce the biases in the model by improving the representation quality of the network as discussed in Sec.~\ref{sec:motivation}. In our approach, we use the SimSiam~\cite{chen2020exploring} self-supervision task as an auxiliary task. We perform ablations in Sec.~\ref{sec:ablchoicess} to validate this choice. SimSiam~\cite{chen2020exploring} is a state-of-the-art self-supervised learning technique. It trains the network to bring two slightly different views of the same image closer in the representation space of the network. The two views denote two different sets of data augmentation techniques applied to the same image.

\begin{figure}[t]
\centering
    \includegraphics[width=0.3\textwidth]{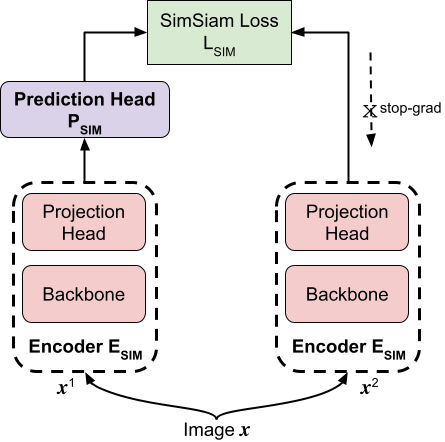}
\caption{Figure shows the SimSiam auxiliary self-supervised task that trains the network to bring two augmented views of the same image close to each other in the feature space. Please note that, at a time the prediction head $P_\texttt{SIM}$ is applied to the encoder output of either the first view $x^1$ or the second view $x^2$ and gradient flow is prevented through the encoder output of the other view \cite{chen2020exploring}.} 
\label{fig:simsiam}
\end{figure}

The two views of the same image are processed by an encoder ($E_{\texttt{SIM}}$) consisting of a backbone network and a multi-layer perceptron projection head. The encoder shares the same parameter weights between the two views. Let $x_i^1,x_i^2$ refer to the two randomly augmented views of the same image $x_i$.
\begin{equation}\label{eq:sim1}
    z_i^2 = E_{\texttt{SIM}}(x_i^2)
\end{equation}
Where, $z_i^2$ refers to the encoder output for $x_i^2$.

A multi-layer perceptron based prediction head ($P_{\texttt{SIM}}$) transforms the encoder output of the first view to match the second view (target view). To achieve this objective SimSiam minimizes the negative cosine similarity between the two views (as shown in Fig.~\ref{fig:simsiam}). The cosine similarity loss is implemented in such a way that the target view encoder output is treated as a constant to prevent gradient flow (stop-gradient) through it \cite{chen2020exploring}.
\begin{equation}\label{eq:sim2}
    p_i^1 = P_{\texttt{SIM}}(E_{\texttt{SIM}}(x_i^1))
\end{equation}
\begin{equation}\label{eq:sim3}
    D(p_i^1,z_i^2) = -\frac{p_i^1}{\|p_i^1\|_2}.\frac{z_i^2}{\|z_i^2\|_2}
\end{equation}
Where, $p_i^1$ refers to the prediction MLP output for $x_i^1$. $\|.\|_2$ refers to $\ell_2$-norm. $D$ refers to the negative cosine similarity loss.

The authors in \cite{chen2020exploring} also perform a similar operation using $P_{\texttt{SIM}}$ to transform the second view to match the first view (target view) in order to achieve a symmetrized loss.
\begin{equation}\label{eq:sim4}
    L_{\texttt{SIM}}(x_i^1,x_i^2) = \frac{1}{2}D(p_i^1,z_i^2) + \frac{1}{2}D(p_i^2,z_i^1)
\end{equation}
Where, $p_i^2 = P_{\texttt{SIM}}(E_{\texttt{SIM}}(x_i^2))$ refers to the prediction MLP output for $x_i^2$. $z_i^1 = E_{\texttt{SIM}}(x_i^1)$ refers to the encoder output for $x_i^1$. $L_{\texttt{SIM}}$ is the SimSiam loss.

In our approach, the entire network, apart from the last classification layer, serves as the backbone network. We add a multi-layer perceptron projection head and a multi-layer perceptron based prediction head to our network for the SimSiam self-supervised auxiliary task.

\subsection{Self Distillation from a Trained Teacher}

\begin{figure}[t]
\centering
    \includegraphics[width=0.3\textwidth]{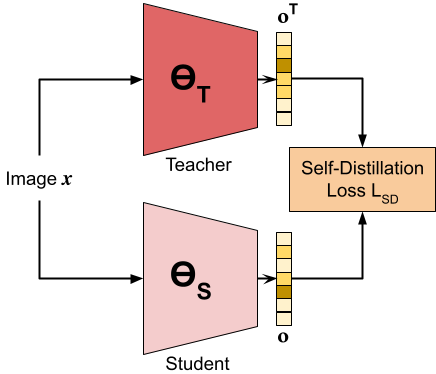}
\caption{Figure shows the self-distillation process from a trained teacher network to the student network. } 
\label{fig:distill}
\end{figure}

The self-distillation process trains a student network that has the same architecture as the trained teacher network. The authors in \cite{NEURIPS2020_2288f691} explain how the self-distillation process improves the generalization of the network. We use the self-distillation process to reduce the impact of biases in the model by improving the generalization power of the network as discussed in Sec.~\ref{sec:motivation}.  The self-distillation process also does not require any additional data making it very suitable for our approach.

In the self-distillation process, we minimize the Kullback–Leibler(KL) divergence between the logits/soft probability outputs of the teacher and student networks (as shown in Fig.~\ref{fig:distill}). This process transfers knowledge from the teacher to the student network and also improves the generalization capacity of the student network. The loss function for self-distillation is defined as follows:

\begin{equation}
    o_i,o_i^{\texttt{T}} = \Theta_{\texttt{S}}(x_i), \Theta_{\texttt{T}}(x_i)
\end{equation}
\begin{equation}
    L_{\texttt{SD}}(o_i,o_i^{\texttt{T}}) =\texttt{KL-Div}(\sigma(o_i),\sigma(o_i^{\texttt{T}}))
\end{equation}

Where, $x_i$ denotes a training data point, $\Theta_{\texttt{S}},\Theta_{\texttt{T}}$ denotes the student and teacher networks, respectively. $o_i,o_i^{\texttt{T}}$ denote the logits produced by the student and teacher network for input $x_i$. $\sigma(.)$ denotes the softmax activation function that transforms the logit $l_c$ for each class $c$ into a probability by comparing it with logits $l_j$ of the remaining classes, i.e., $\sigma(l_c) = \frac{exp^{l_c/\kappa}}{\sum_j exp^{l_j/\kappa}}$. $\kappa$ denotes the temperature hyper-parameter~\cite{hinton2015distill}. $\texttt{KL-Div}$ refers to the KL divergence. $L_{\texttt{SD}}$ refers to the self distillation loss.

In our proposed limited data bias mitigation approach (LDBM), we train a teacher network $\Theta_{\texttt{T}}$ using the cross-entropy loss for the standard classification task and the SimSiam auxiliary self-supervised task loss. The loss function for training the teacher network is defined as follows:
\begin{equation}
    L_{\texttt{T}} = \frac{\Sigma_{i=1}^{N} \{L_{\texttt{CE}}(x_i,y_i) + L_{\texttt{SIM}}(x_i^1,x_i^2)\}}{N}
\end{equation}
Where, $L_{\texttt{CE}}$ represents the cross-entropy loss, $L_{\texttt{SIM}}$ represents the SimSiam loss, $N$ refers to the number of training samples, $y_i$ refers to the label of image $x_i$, and $L_{\texttt{T}}$ refers to the total loss for the teacher network.

Next, we train a student network $\Theta_{\texttt{S}}$ that has the same architecture as the teacher network $\Theta_{\texttt{T}}$ using the cross-entropy loss for the standard classification task, the SimSiam auxiliary self-supervised task loss, and the self distillation loss from the trained teacher network $\Theta_{\texttt{T}}$. For the self-distillation loss, we minimize the KL divergence between soft predictions of $\Theta_{\texttt{S}}$ and $\Theta_{\texttt{T}}$ for the same input image. The total loss $L_{\texttt{S}}$ for training the student network is defined as follows:
\begin{equation}
    L_{\texttt{S}} = \frac{\Sigma_{i=1}^{N}\{L_{\texttt{CE}}(x_i,y_i) + L_{\texttt{SIM}}(x_i^1,x_i^2) + L_{\texttt{SD}}(o_i,o_i^{\texttt{T}})\}}{N}
\end{equation}

After the training is completed, the resulting student network is our final model.

\section{Experiments}
\subsection{Datasets}\label{sec:dataset}
We perform the experiments on the limited L-CIFAR-10S skewed dataset described in Sec.~\ref{sec:limiteddata}, which is created from the CIFAR-10S dataset~\cite{wang2020towards} with the protected attribute/domain being the color/grayscale image. The authors in \cite{wang2020towards} also create modified versions of CIFAR-10S by changing the domain attribute. The new domains consist of normal CIFAR-10 images and either a) CIFAR-10S-i: images of the same class from the ImageNet~\cite{russakovsky2015imagenet} dataset (CIFAR-10S-i) downsampled to $32\times32$, or, b) CIFAR-10S-c28: images cropped to $28\times28$ and upsampled to $32\times32$, or, c) CIFAR-10S-d16: images downsampled to $16\times16$ and upsampled to $32\times32$, or d) CIFAR-10S-d8: images downsampled to $8\times8$ and upsampled to $32\times32$. All these datasets have the same bias problem as in CIFAR-10S and L-CIFAR-10S. We create the limited data versions of all these datasets, i.e., L-CIFAR-10S-i, L-CIFAR-10S-c28, L-CIFAR-10S-d16, and L-CIFAR-10S-d8, using the same approach as described in Sec.~\ref{sec:limiteddata} to reduce the training data to 5\% without changing the level of skew. We perform experiments using the 95-5\% skew (described in Sec.~\ref{sec:fullstory}) for all these datasets. We also perform experiments on the CelebA multi-labeled dataset after reducing the training set to the first 5\% images (L-CelebA). Further details regarding these datasets have been provided in the appendix.

\subsection{Compared Approaches}\label{sec:compappr}
We perform experiments on the naive baseline method in the limited data setting. It trains using the cross-entropy loss without applying any bias mitigation approaches. We also perform experiments on the strategic sampling, adversarial, domain discriminative, and domain independent approaches described in \cite{wang2020towards}. The strategic sampling over samples the minority image type in each class, e.g., color images in grayscale domain classes and vice-versa. The adversarial method minimizes the ability of the network to identify the protected attribute/domain, e.g., whether the image is colored or grayscale. The domain discriminative method trains an ND-way classifier where N is the number of image classes and D is the number of domains (values that the protected attribute can take). The domain independent method trains separate N-way classifiers for each domain D. We have provided details regarding these approaches in the appendix. 

\subsection{Implementation Details} 
In order to use SimSiam as an auxiliary task in our approach, we use the ResNet-18 architecture as the backbone network for the L-CIFAR-10S experiments and ResNet-50 architecture (pretrained on ImageNet) as the backbone network for the L-CelebA experiments. Specifically, we take the output of the last convolutional layer before the fully connected classification layer and use it for the SimSiam objective function. We add a multi-layer perceptron projection head and a multi-layer perceptron based prediction head as required by SimSiam. The multi-layer perceptron projection head has 3 fully connected layers, each having an output size of 2048 and followed by a batch normalization layer and ReLU activation. The multi-layer perceptron prediction head has 2 fully connected layers having output sizes of 512 and 2048, respectively. The first layer is followed by a batch normalization layer and ReLU activation. The training settings for SimSiam are the same as proposed in \cite{chen2020exploring}. For self-distillation, we use $\kappa=4$. For a fair comparison, we use the same training settings for all the methods as described in \cite{wang2020towards}. For details regarding the training settings refer to the appendix. 

For the L-CIFAR-10S experiments, we use two metrics also used by \cite{wang2020towards}, i.e., mean per-class per-domain accuracy (primary) and bias amplification \cite{zhao_men_2017}. Since the overall test set is fully balanced across the two types of image in each version of L-CIFAR-10S, the mean accuracy reflects the effect of biases on the predictions of the model \cite{wang2020towards}. For better understanding and completeness w.r.t. to \cite{wang2020towards}, we also provide the mean bias score, defined as follows:
\begin{equation}
\label{eq:bias}
\frac{1}{|C|}\sum_{c \in C} \frac{\max(\mathrm{Gr}_c, \mathrm{Col}_c)}{\mathrm{Gr}_c + \mathrm{Col}_c} - 0.5.
\end{equation}

Where, $\mathrm{Gr}_c,\mathrm{Col}_c$ refer to the number of grayscale and color images predicted to belong to the class $c\in C$.

For the L-CelebA experiments, we perform multi-label classification that generally uses the metric of mean average precision (mAP) across attributes. We use a weighted mAP proposed in \cite{wang2020towards} to remove the gender bias in the test set. Following \cite{wang2020towards}, we also provide the bias amplification (BA) over the attributes~\cite{zhao_men_2017}. For further details, refer to the appendix. 

\subsection{L-CIFAR-10S Results}
\begin{table}[t]
\centering
\begin{footnotesize}
\scalebox{0.69}{
\addtolength{\tabcolsep}{-5pt}
\begin{tabular}{l?l?l?c?ccc}
\toprule
&&&& \multicolumn{3}{c}{\ Accuracy (\%, $\uparrow$)}\\
{Model Name} & {Model} & {Test Inference} & {Bias ($\downarrow$)} & {Color} & {Gray} & {Mean} \\
\midrule
{Baseline} & N-way  & $\argmax_y\mathrm{P}(y|x)$ & $0.349$ & $41.04$ & $42.11$ & $41.58 \pm 0.6$\\
\cmidrule{2-7}
\textbf{w/ LDBM (Ours)} & N-way  & $\argmax_y\mathrm{P}(y|x)$ & $0.076$ & $53.31$ & $47.63$ & $50.47 \pm 0.7$\\
\midrule
{Sampling} & N-way  & $\argmax_y\mathrm{P}(y|x)$ & $0.265$ & $50.96$ & $52.43$ & $51.7\pm0.8$\\
\cmidrule{2-7}
\textbf{w/ LDBM (Ours)} & N-way  & $\argmax_y\mathrm{P}(y|x)$ & $0.027$ & $56.96$ & $57.15$ & $57.01\pm0.7$\\
\midrule
\multirow{2}{*}{{Adversarial}} & w/ unf. conf.  & $\argmax_y\mathrm{P}(y|x)$ & $0.362$ &
$40.98$ & $44.24$ & $42.61\pm1.4$ \\
& w/ \(\nabla\) rev., proj.  & $\argmax_y\mathrm{P}(y|x)$ & $0.324$ &
$40.19$ & $45.09$ & $42.64\pm1.4$\\
\cmidrule{2-7}
\textbf{w/ LDBM (Ours)} & w/ unf. conf.  & $\argmax_y\mathrm{P}(y|x)$ & $0.065$ &
$54.42$ & $50.99$ & $52.71\pm1.3$ \\
\midrule 
\multirow{4}{*}{{DomainDiscrim}} & \multirow{3}{*}{joint ND-way }  
&  $\argmax_y\sum_d \ptr(y,d|x)$ & $0.390$&$38.05$ & $36.72$ & $37.39\pm0.7$ \\ 
& & $\argmax_y\max_d \pte(y,d|x)$ & $0.232$ &$41.38$ & $41.3$ & $41.34\pm0.9$ \\ 
& & $\argmax_y\sum_d \pte(y,d|x)$ & $0.232$&
$41.47$ & $41.33$ & $41.4\pm0.8$ \\
\cmidrule{2-7}
& RBA \cite{zhao_men_2017} & y=\(\Lagr(\sum_d \ptr(y, d | x))\) & $0.252$ & $38.21$ & $38.23$ & $38.22\pm0.9$ \\
\cmidrule{2-7}
\textbf{w/ LDBM (Ours)} & \multirow{1}{*}{joint ND-way }  
&  $\argmax_y\sum_d \pte(y,d|x)$ & $0.079$&
$57.75$ & $53.81$ & $55.78\pm0.7$ \\
\midrule
\multirow{2}{*}{{DomInd}} & \multirow{2}{*}{N-way per dom.} & 
 $\argmax_y\pte(y|\knownd, x)$ & $0.319$ & $50.83$ & $46.25$ & $48.54\pm0.6$ \\
&&$\argmax_y\sum_d s(y,d,x)$ & $0.027$ &$59.25$ & $58.39$ & $58.82\pm0.4$ \\
\cmidrule{2-7}
\multirow{2}{*}{\textbf{w/ LDBM (Ours)}} & \multirow{2}{*}{N-way per dom.} & 
 $\argmax_y\pte(y|\knownd, x)$ & $0.333$ & $53.9$ & $50.92$ & $52.41\pm0.5$ \\
&&$\argmax_y\sum_d s(y,d,x)$ & $\mathbf{0.011}$ &$\mathbf{66.42}$ & $\mathbf{65.37}$ & $\mathbf{65.90\pm0.6}$ \\
\bottomrule
\end{tabular}
}
\end{footnotesize}
\caption{Performance comparison of bias mitigation strategies on L-CIFAR-10S using the ResNet-18 \cite{he_deep_2015}. }
\label{table:cifar}
\end{table}

\begin{table}[t]
\centering
\begin{footnotesize}
\scalebox{0.69}{
\addtolength{\tabcolsep}{-5pt}
\begin{tabular}{l?l?l?c?ccc}
\toprule
&&&& \multicolumn{3}{c}{\ Accuracy (\%, $\uparrow$)}\\
{Model Name} & {Model} & {Test Inference} & {Bias ($\downarrow$)} & {Color} & {ImgNet} & {Mean} \\
\midrule
{Baseline} & N-way  & $\argmax_y\mathrm{P}(y|x)$ & $0.289$ & $47.94$ & $36.5$ & $42.22 \pm 0.5$\\
\cmidrule{2-7}
\textbf{w/ LDBM (Ours)} & N-way  & $\argmax_y\mathrm{P}(y|x)$ & $0.091$ & $66.56$ & $49.27$ & $57.92 \pm 0.4$\\
\midrule
{Sampling} & N-way  & $\argmax_y\mathrm{P}(y|x)$ & $0.208$ & $50.16$ & $40.11$ & $45.14\pm0.5$\\
\midrule
\multirow{2}{*}{{Adversarial}} & w/ unf. conf.  & $\argmax_y\mathrm{P}(y|x)$ & $0.257$ &
$50.32$ & $39.28$ & $44.8\pm1.2$ \\
& w/ \(\nabla\) rev., proj.  & $\argmax_y\mathrm{P}(y|x)$ & $0.243$ &
$49.76$ & $38.29$ & $44.03\pm1.3$\\
\midrule 
\multirow{4}{*}{{DomainDiscrim}} & \multirow{3}{*}{joint ND-way }  
&  $\argmax_y\sum_d \ptr(y,d|x)$ & $0.330$&$48.5$ & $36.79$ & $42.64\pm0.7$ \\ 
& & $\argmax_y\max_d \pte(y,d|x)$ & $0.203$ &$51.94$ & $39.64$ & $45.79\pm0.8$ \\ 
& & $\argmax_y\sum_d \pte(y,d|x)$ & $0.203$&
$52.12$ & $39.73$ & $45.92\pm0.6$ \\
\cmidrule{2-7}
& RBA \cite{zhao_men_2017} & y=\(\Lagr(\sum_d \ptr(y, d | x))\) & $0.213$ & $50.88$ & $37.68$ & $44.28\pm0.6$ \\
\midrule
\multirow{2}{*}{{DomInd}} & \multirow{2}{*}{N-way per dom.} & 
 $\argmax_y\pte(y|\knownd, x)$ & $0.381$ & $42.62$ & $34.30$ & $38.46\pm0.7$ \\
&&$\argmax_y\sum_d s(y,d,x)$ & $0.038$ &$51.11$ & $42.18$ & $46.64\pm0.5$ \\
\cmidrule{2-7}
\multirow{2}{*}{\textbf{w/ LDBM (Ours)}} & \multirow{2}{*}{N-way per dom.} & 
 $\argmax_y\pte(y|\knownd, x)$ & $0.364$ & $51.23$ & $39.24$ & $45.24\pm0.6$ \\
&&$\argmax_y\sum_d s(y,d,x)$ & $\mathbf{0.022}$ &$\mathbf{63.55}$ & $\mathbf{47.93}$ & $\mathbf{55.74\pm0.6}$ \\
\bottomrule
\end{tabular}
}
\end{footnotesize}
\caption{Performance comparison of bias mitigation strategies on L-CIFAR-10S-i using the ResNet-18 \cite{he_deep_2015}. }
\label{table:cifarimgnet}
\end{table}

\begin{table}[t]
\centering
\begin{footnotesize}
\scalebox{0.69}{
\addtolength{\tabcolsep}{-5pt}
\begin{tabular}{l?l?l?c?ccc}
\toprule
&&&& \multicolumn{3}{c}{\ Accuracy (\%, $\uparrow$)}\\
{Model Name} & {Model} & {Test Inference} & {Bias ($\downarrow$)} & {Color} & {Crop} & {Mean} \\
\midrule
{Baseline} & N-way  & $\argmax_y\mathrm{P}(y|x)$ & $0.025$ & $42.47$ & $41.93$ & $42.2 \pm 0.6$\\
\cmidrule{2-7}
\textbf{w/ LDBM (Ours)} & N-way  & $\argmax_y\mathrm{P}(y|x)$ & $0.016$ & $51.48$ & $53.31$ & $52.39 \pm 0.4$\\
\midrule
{Sampling} & N-way  & $\argmax_y\mathrm{P}(y|x)$ & $0.032$ & $47.75$ & $47.46$ & $47.61\pm0.6$\\
\midrule
\multirow{2}{*}{{Adversarial}} & w/ unf. conf.  & $\argmax_y\mathrm{P}(y|x)$ & $0.037$ &
$44.87$ & $43.78$ & $44.325\pm1.2$ \\
& w/ \(\nabla\) rev., proj.  & $\argmax_y\mathrm{P}(y|x)$ & $0.029$ &
$45.95$ & $43.96$ & $44.96\pm1.1$\\
\midrule 
\multirow{4}{*}{{DomainDiscrim}} & \multirow{3}{*}{joint ND-way }  
&  $\argmax_y\sum_d \ptr(y,d|x)$ & $0.024$&$39.55$ & $38.67$ & $39.11\pm0.5$ \\ 
& & $\argmax_y\max_d \pte(y,d|x)$ & $0.025$ &$42.71$ & $42.05$ & 42.38$\pm0.6$ \\ 
& & $\argmax_y\sum_d \pte(y,d|x)$ & $0.025$&
$42.78$ & $42.08$ & $42.43\pm0.8$ \\
\cmidrule{2-7}
& RBA \cite{zhao_men_2017} & y=\(\Lagr(\sum_d \ptr(y, d | x))\) & $0.023$ & $39.58$ & $38.67$ & $39.13\pm0.6$ \\
\midrule
\multirow{2}{*}{{DomInd}} & \multirow{2}{*}{N-way per dom.} & 
 $\argmax_y\pte(y|\knownd, x)$ & $0.336$ & $47.87$ & $41.59$ & $44.73\pm0.4$ \\
&&$\argmax_y\sum_d s(y,d,x)$ & $0.035$ &$56.34$ & $55.39$ & $55.86\pm0.3$ \\
\cmidrule{2-7}
\multirow{2}{*}{\textbf{w/ LDBM (Ours)}} & \multirow{2}{*}{N-way per dom.} & 
 $\argmax_y\pte(y|\knownd, x)$ & $0.322$ & $53.63$ & $53.1$ & $53.37\pm0.6$ \\
&&$\argmax_y\sum_d s(y,d,x)$ & $\mathbf{0.013}$ &$\mathbf{65.14}$ & $\mathbf{67.5}$ & $\mathbf{66.32\pm0.4}$ \\
\bottomrule
\end{tabular}
}
\end{footnotesize}
\caption{Performance comparison of bias mitigation strategies on L-CIFAR-10S-c28 using the ResNet-18 \cite{he_deep_2015}. }
\label{table:cifarcrop}
\end{table}

\begin{table}[t]
\centering
\begin{footnotesize}
\scalebox{0.69}{
\addtolength{\tabcolsep}{-5pt}
\begin{tabular}{l?l?l?c?ccc}
\toprule
&&&& \multicolumn{3}{c}{\ Accuracy (\%, $\uparrow$)}\\
{Model Name} & {Model} & {Test Inference} & {Bias ($\downarrow$)} & {Color} & {16x16} & {Mean} \\
\midrule
{Baseline} & N-way  & $\argmax_y\mathrm{P}(y|x)$ & $0.146$ & $44.79$ & $36.61$ & $40.7 \pm 0.7$\\
\cmidrule{2-7}
\textbf{w/ LDBM (Ours)} & N-way  & $\argmax_y\mathrm{P}(y|x)$ & $0.065$ & $49.26$ & $46.89$ & $48.075 \pm 0.8$\\
\midrule
{Sampling} & N-way  & $\argmax_y\mathrm{P}(y|x)$ & $0.167$ & $44.84$ & $41.38$ & $43.11\pm0.6$\\
\midrule
\multirow{2}{*}{{Adversarial}} & w/ unf. conf.  & $\argmax_y\mathrm{P}(y|x)$ & $0.136$ &
$44.42$ & $38.0$ & $41.21\pm1.1$ \\
& w/ \(\nabla\) rev., proj.  & $\argmax_y\mathrm{P}(y|x)$ & $0.146$ &
$44.18$ & $38.54$ & $41.36\pm1.4$\\
\midrule 
\multirow{4}{*}{{DomainDiscrim}} & \multirow{3}{*}{joint ND-way }  
&  $\argmax_y\sum_d \ptr(y,d|x)$ & $0.123$&$43.33$ & $33.99$ & $38.66\pm0.7$ \\ 
& & $\argmax_y\max_d \pte(y,d|x)$ & $0.102$ &$46.87$ & $35.92$ & 41.39$\pm0.6$ \\ 
& & $\argmax_y\sum_d \pte(y,d|x)$ & $0.103$&
$46.94$ & $36.08$ & $41.51\pm0.6$ \\
\cmidrule{2-7}
& RBA \cite{zhao_men_2017} & y=\(\Lagr(\sum_d \ptr(y, d | x))\) & $0.067$ & $43.37$ & $33.99$ & $38.68\pm0.7$ \\
\midrule
\multirow{2}{*}{{DomInd}} & \multirow{2}{*}{N-way per dom.} & 
 $\argmax_y\pte(y|\knownd, x)$ & $0.366$ & $45.51$ & $35.27$ & $40.39\pm0.7$ \\
&&$\argmax_y\sum_d s(y,d,x)$ & $0.088$ &$56.34$ & $55.39$ & $51.5\pm0.6$ \\
\cmidrule{2-7}
\multirow{2}{*}{\textbf{w/ LDBM (Ours)}} & \multirow{2}{*}{N-way per dom.} & 
 $\argmax_y\pte(y|\knownd, x)$ & $0.342$ & $49.93$ & $47.85$ & $48.89\pm0.8$ \\
&&$\argmax_y\sum_d s(y,d,x)$ & $\mathbf{0.026}$ &$\mathbf{63.18}$ & $\mathbf{59.8}$ & $\mathbf{61.49\pm0.6}$ \\
\bottomrule
\end{tabular}
}
\end{footnotesize}
\caption{Performance comparison of bias mitigation strategies on L-CIFAR-10S-d16 using the ResNet-18 \cite{he_deep_2015}. }
\label{table:cifard16}
\end{table}

\begin{table}[t]
\centering
\begin{footnotesize}
\scalebox{0.69}{
\addtolength{\tabcolsep}{-5pt}
\begin{tabular}{l?l?l?c?ccc}
\toprule
&&&& \multicolumn{3}{c}{\ Accuracy (\%, $\uparrow$)}\\
{Model Name} & {Model} & {Test Inference} & {Bias ($\downarrow$)} & {Color} & {8x8} & {Mean} \\
\midrule
{Baseline} & N-way  & $\argmax_y\mathrm{P}(y|x)$ & $0.242$ & $41.03$ & $27.74$ & $34.39 \pm 0.7$\\
\cmidrule{2-7}
\textbf{w/ LDBM (Ours)} & N-way  & $\argmax_y\mathrm{P}(y|x)$ & $0.175$ & $47.89$ & $33.37$ & $40.63 \pm 0.6$\\
\midrule
{Sampling} & N-way  & $\argmax_y\mathrm{P}(y|x)$ & $0.237$ & $40.09$ & $32.78$ & $36.44\pm0.8$\\
\midrule
\multirow{2}{*}{{Adversarial}} & w/ unf. conf.  & $\argmax_y\mathrm{P}(y|x)$ & $0.220$ &
$41.05$ & $27.21$ & $34.13\pm1.4$ \\
& w/ \(\nabla\) rev., proj.  & $\argmax_y\mathrm{P}(y|x)$ & $0.237$ &
$41.13$ & $25.49$ & $33.31\pm1.3$\\
\midrule 
\multirow{4}{*}{{DomainDiscrim}} & \multirow{3}{*}{joint ND-way }  
&  $\argmax_y\sum_d \ptr(y,d|x)$ & $0.219$&$39.17$ & $25.75$ & $32.46\pm0.7$ \\ 
& & $\argmax_y\max_d \pte(y,d|x)$ & $0.137$ &$42.81$ & $27.28$ & 35.04$\pm0.4$ \\ 
& & $\argmax_y\sum_d \pte(y,d|x)$ & $0.136$&
$42.79$ & $27.41$ & $35.1\pm0.5$ \\
\cmidrule{2-7}
& RBA \cite{zhao_men_2017} & y=\(\Lagr(\sum_d \ptr(y, d | x))\) & $0.133$ & $39.25$ & $25.78$ & $32.52\pm0.7$ \\
\midrule
\multirow{2}{*}{{DomInd}} & \multirow{2}{*}{N-way per dom.} & 
 $\argmax_y\pte(y|\knownd, x)$ & $0.411$ & $41.62$ & $28.48$ & $35.05\pm0.6$ \\
&&$\argmax_y\sum_d s(y,d,x)$ & $0.110$ &$49.09$ & $37.64$ & $43.37\pm0.3$ \\
\cmidrule{2-7}
\multirow{2}{*}{\textbf{w/ LDBM (Ours)}} & \multirow{2}{*}{N-way per dom.} & 
 $\argmax_y\pte(y|\knownd, x)$ & $0.403$ & $47.11$ & $33.1$ & $40.11\pm0.7$ \\
&&$\argmax_y\sum_d s(y,d,x)$ & $\mathbf{0.103}$ &$\mathbf{56.19}$ & $\mathbf{43.17}$ & $\mathbf{49.68\pm0.4}$ \\
\bottomrule
\end{tabular}
}
\end{footnotesize}
\caption{Performance comparison of bias mitigation strategies on L-CIFAR-10S-d8 using the ResNet-18 \cite{he_deep_2015}. }
\label{table:cifard8}
\end{table}

Tables~\ref{table:cifar}, \ref{table:cifarimgnet}, \ref{table:cifarcrop}, \ref{table:cifard16}, \ref{table:cifard8} compare the performance of different bias mitigation strategies including our proposed LDBM approach on L-CIFAR-10S based datasets. The results indicate that the performance of the baselines and all bias mitigation strategies have been significantly affected in the limited data setting (as compared to the full data results reported in \cite{wang2020towards}, Fig.~\ref{fig:fulldata}). We observe that our proposed LDBM approach significantly improves the performance of the baseline model for all the datasets. Specifically, for L-CIFAR-10S with 95-5\% skew, LDBM significantly reduces the bias score of the baseline model by 78.22\% and improves its performance by a significant absolute margin of 8.89\% in the mean accuracy. LDBM also significantly reduces the bias score for the state-of-the-art domain independent bias mitigation method by 59.26\% and improves its performance by a significant absolute margin of 7.08\% in the mean accuracy. In fact, the domain independent training approach with our proposed LDBM achieves a slightly better performance on the color test set than even the model trained on all grayscale version of L-CIFAR-10S (GTCT 65.57\%) and very close to the performance of the all color version of L-CIFAR-10S (CTCT 66.53\%) as described in Sec.~\ref{sec:story}. LDBM also significantly reduces the bias scores for the strategic sampling, adversarial, and domain discriminative bias mitigation approaches by 89.81\%, 82.04\%, 65.95\%, respectively, and significantly improves their performance by absolute margins of 5.31\%, 10.1\%, and 14.38\%, respectively, on L-CIFAR-10S.

For the other L-CIFAR-10S variants, we use our approach to augment the state-of-the-art domain independent bias mitigation technique. We observe in Table~\ref{table:cifarimgnet} that for the L-CIFAR-10S-i dataset, LDBM significantly reduces the bias score of the baseline model by 68.51\% and improves its performance by a significant absolute margin of 15.7\%. It also significantly reduces the bias score of the domain independent bias mitigation method by 42.11\% and improves its performance by a significant absolute margin of 9.1\%. Therefore, our proposed LDBM significantly reduces the effect of biases on the model predictions and also significantly improves the performance of bias mitigation strategies.

\subsection{L-CelebA Results}

\begin{table}[t]
    \centering
    \footnotesize
    \scalebox{0.8}{
    \addtolength{\tabcolsep}{3pt}
    \begin{tabular}{l?l?c?c}
\toprule
{Model Name} & {Model} & {mAP ($\uparrow$)} & {BA ($\downarrow$)} \\
\hline
{Base}  &  N sigmoids & 65.34 & 0.036\\
\cmidrule{2-4}
{\textbf{w/ LDBM (Ours)}}  &  N sigmoids & 66.72 & 0.017\\
\midrule
{Adversarial} 
&  w/uniform conf.~\cite{alvi2018turning,tzeng2015simultaneous}

& 56.82 & 0.070\\

\multirow{1}{*}{{DomainDiscrim}} & 2N sigm, $\sum_d \mathrm{P_{tr}}(y,d|x)$ & 64.71 & 0.020\\ 
\midrule
\multirow{4}{*}{{DomInd}} & 2N sigmoids, $\mathrm{P_{tr}}(y|d^*,x)$ & 63.77 & 0.027\\
& 2N sigm, $\max_d \mathrm{P_{tr}}(y|d,x)$ & 64.62 & {-0.048}\\
& 2N sigm, $\sum_d \mathrm{P_{tr}}(y|d,x)$ & 65.35 & -0.006\\
& 2N sigmoids, $\sum_d s(y,d,x)$ & { 65.73} & -0.007\\
\cmidrule{2-4}
\multirow{4}{*}{{\textbf{w/ LDBM (Ours)}}} & 2N sigmoids, $\mathrm{P_{tr}}(y|d^*,x)$ & 66.36 & 0.024\\
& 2N sigm, $\max_d \mathrm{P_{tr}}(y|d,x)$ & 67.92 & {-0.051}\\
& 2N sigm, $\sum_d \mathrm{P_{tr}}(y|d,x)$ & 68.49 & \bf  -0.073\\
& 2N sigmoids, $\sum_d s(y,d,x)$ & {\bf 69.07} & -0.019\\
\bottomrule
    \end{tabular}
    }
    \caption{Performance of bias mitigation approaches on the L-CelebA dataset using ResNet-50.}
    \label{table:celeba}
\end{table}

Tables~\ref{table:celeba} compares the performance of different bias mitigation strategies, including our proposed LDBM approach on the L-CelebA dataset. We observe that the effectiveness of the bias mitigation strategies diminishes as compared to the full data results reported in \cite{wang2020towards}. We also observe that our proposed LDBM approach significantly improves the performance of the baseline approach and the domain-independent training approach with respect to average mAP across attributes and the bias amplification score.

\section{Ablation Experiments}

\subsection{Bias Amplification Score over Training Epochs}\label{sec:ablgeneral}
\begin{figure}[t]
    \centering
     \includegraphics[width=0.4\textwidth]{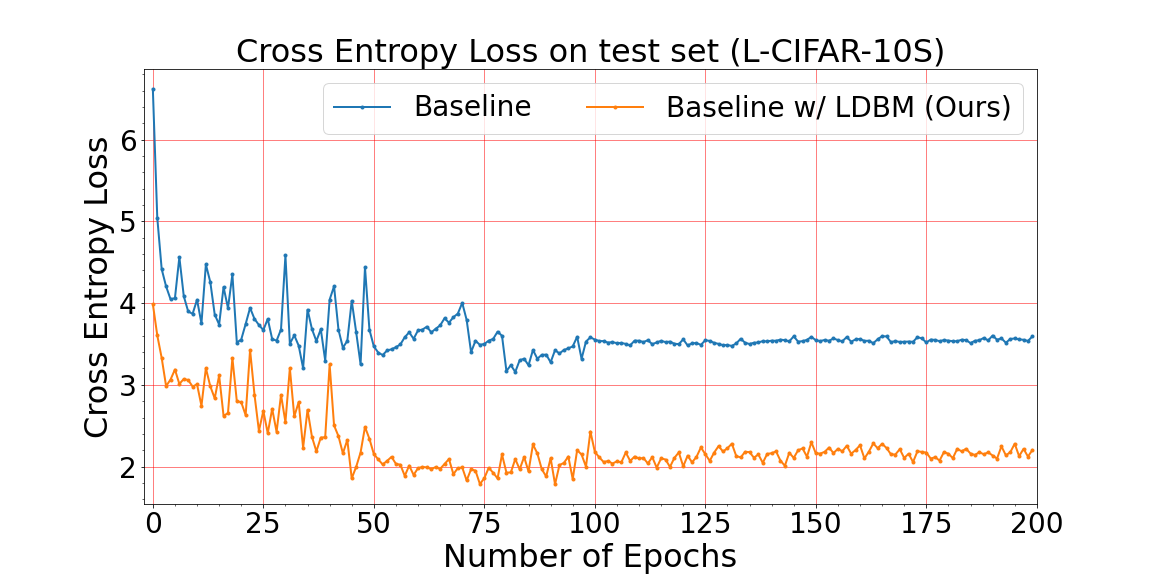}
    \caption{Experimental results showing the test loss (cross-entropy loss) of the network for baseline with or without LDBM trained on L-CIFAR-10S using ResNet-18.}\label{fig:generalize}
\end{figure}

\begin{figure}[t]
    \centering
     \includegraphics[width=0.4\textwidth]{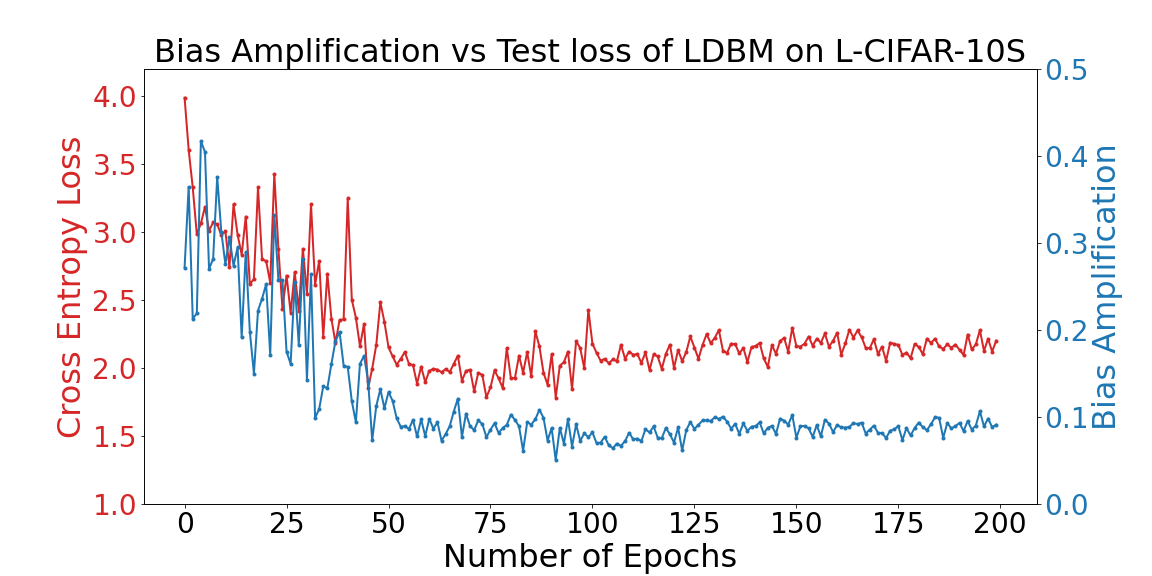}
    \caption{Experimental results showing the decrease in bias amplification scores with the decrease in test loss (improved generalization) of the network when baseline w/ LDBM is trained on L-CIFAR-10S using ResNet-18.}\label{fig:generalizebias}
\end{figure}

Fig.~\ref{fig:generalize} compares the cross-entropy loss on the L-CIFAR-10S test set using the baseline approach and baseline w/ LDBM approach across the training epochs. The results indicate that the baseline w/ LDBM encounters lower test loss than the baseline model. Therefore, the generalization power and representation quality of the model obtained using our approach is better than that of the baseline model. Fig.~\ref{fig:generalizebias} shows the decrease in the bias amplification score as the model gets trained. Figs.~\ref{fig:generalize}, \ref{fig:generalizebias} indicate that our LDBM approach reduces the bias score of the model by improving the generalization ability and representation quality of the model.

\subsection{Significance of Components}

\begin{table}[t]
    \centering
    \footnotesize
    \scalebox{0.8}{
    \addtolength{\tabcolsep}{-3pt}
    \begin{tabular}{cccccc}
\toprule
 {Self Sup.} & {Self-Distill} & {Bias ($\downarrow$)} & {Color Acc. \%,($\downarrow$)} & {Gray Acc. \%,($\uparrow$)}  & {Mean Acc. \%,($\uparrow$)} \\
\hline
\xmark &  \xmark  & 0.027 &59.25 & 58.39 & 58.82$\pm$0.7\\
\xmark  &  \cmark  & 0.024 & 61.23 & 58.96 & 60.10$\pm$0.6\\
\cmark  &  \xmark  & 0.013 & 65.20 & 64.24 & 64.72$\pm$0.5\\
\cmark  &  \cmark & \textbf{0.011} &\textbf{66.42} & \textbf{65.37} & \textbf{65.90$\pm$0.6}\\
\bottomrule
    \end{tabular}
    }
    \caption{Ablation experiment showing the significance of different components of our approach applied to the domain independent method on L-CIFAR-10S.}
    \label{table:ablcomp}
\end{table}

The results in Table \ref{table:ablcomp} indicate that both the auxiliary self-supervision and the self-distillation components of our approach are essential to the performance of our approach.

\subsection{Choice of Self-Supervision}\label{sec:ablchoicess}
\begin{table}[t]
    \centering
    \footnotesize
    \scalebox{0.8}{
    \addtolength{\tabcolsep}{-1pt}
    \begin{tabular}{ccccc}
\toprule
 {Self Supervision} & {Bias ($\downarrow$)} & {Color Acc. \%,($\downarrow$)} & {Gray Acc. \%,($\uparrow$)}  & {Mean Acc. \%,($\uparrow$)} \\
\hline
Rotation & 0.026 & 59.85 & 59.41 & 59.63$\pm$0.8\\
SimClr & 0.017 & 62.81 & 62.03 & 62.42$\pm$0.5\\
\textbf{SimSiam} &  \textbf{0.011} & \textbf{66.42} & \textbf{65.37} & \textbf{65.90$\pm$0.6}\\
\bottomrule
    \end{tabular}
    }
    \caption{Ablation experiment showing the performance of different self-supervision techniques used as an auxiliary task along with self-distillation in our approach LDBM applied to the domain independent method on L-CIFAR-10S.}
    \label{table:ablss}
\end{table}

The results in Table \ref{table:ablss} indicate that the SimSiam based self-supervised auxiliary task is more effective in our approach as compared to other techniques. The contrastive training approach in the SimSiam and SimClr techniques helps the network to learn better features from the images.

\subsection{Skew Level vs. Accuracy}
\begin{figure}[t]
    \centering
     \includegraphics[width=0.45\textwidth]{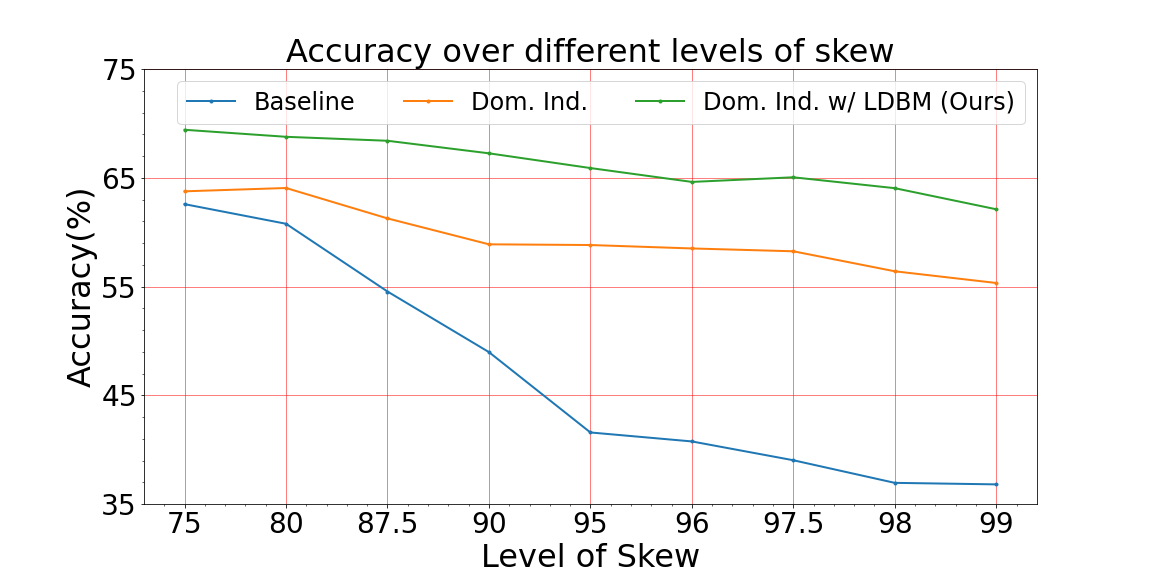}
    \caption{Figure shows the performance of the baseline, domain independent and domain independent w/ LDBM methods for different levels of skew in L-CIFAR-10S.}\label{fig:skew}
\end{figure}

The results in Fig.~\ref{fig:skew} indicate that as the level of skew increases, the performance of the baseline decreases significantly due to an increase in the bias in the data. However, the performance of our approach remains relatively stable and significantly higher than the others.

\subsection{LDBM complements Bias Mitigation Methods}

The results in Table~\ref{table:cifar} indicate that our proposed LDBM can complement other bias mitigation approaches, and significantly reduces the bias scores of these approaches in the limited data setting. Specifically, it significantly reduces the bias score for the strategic sampling, adversarial, domain discriminative, and domain independent bias mitigation approaches by 89.81\%, 82.04\%, 65.95\%, and 59.26\%, respectively on L-CIFAR-10S. It also significantly improves their performance in the limited data setting. This makes our approach very useful and effective.

\subsection{Performance improvement on CIFAR-10S}\label{sec:performsufdata}

\begin{figure}[t]
    \centering
     \includegraphics[width=0.4\textwidth]{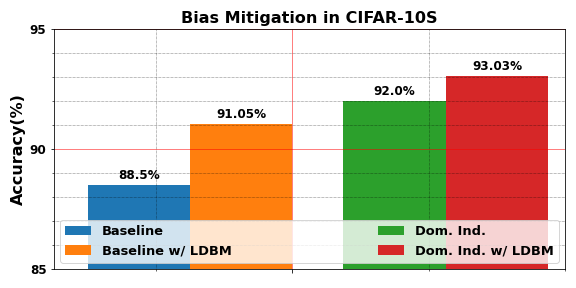}
    \caption{Figure shows the performance (mean of color \& grayscale accuracy) of the baseline, baseline w/ LDBM, domain independent and domain independent w/ LDBM methods on the CIFAR-10S dataset.}\label{fig:fulldata}
\end{figure}

The results in Fig.~\ref{fig:fulldata} indicate that our proposed approach outperforms the baseline method by an absolute margin of 2.55\% on the CIFAR-10S dataset, which has a sufficient number of training examples per class. The state-of-the-art domain independent training approach augmented with our proposed LDBM achieves around 1\% improvement as compared to 7.08\% in the limited data setting. In this setting, since the network is trained on a sufficient amount of data, it is less prone to learning irrelevant features and unwanted correlations from the training data. However, our approach still improves the performance of the network by further reducing the impact of the biases. In fact, the domain independent method w/ LDBM (93.2\%) slightly outperforms GTCT (93.0\%) on the color test set.

\section{Conclusion}
In this paper, we showed that when the number of training examples is limited, the bias in the dataset has a severe impact on the performance of the model. We also experimentally demonstrated that bias mitigation approaches become less effective with limited training data, which is a novel finding. We proposed a novel approach for reducing the impact of such biases on the model predictions. Specifically, we demonstrated how self-supervision and self-distillation, which are not used for bias mitigation, can be adapted to significantly reduce the impact of biases on the model, especially in the limited data regime, which is an interesting finding. We empirically demonstrated how our proposed limited data bias mitigation (LDBM) approach significantly reduces the bias score of the baseline model and significantly improves its performance. We also demonstrated that our approach complements other bias mitigation approaches. It significantly reduces the bias score of these approaches and significantly improves their performance.

{\small
\bibliographystyle{ieee_fullname}
\bibliography{egbib}

\begin{thebibliography}{10}\itemsep=-1pt

\bibitem{alvi2018turning}
Mohsan Alvi, Andrew Zisserman, and Christoffer Nellaker.
\newblock Turning a blind eye: Explicit removal of biases and variation from
  deep neural network embeddings.
\newblock In {\em ECCV}, 2018.

\bibitem{anne2018women}
Lisa Anne~Hendricks, Kaylee Burns, Kate Saenko, Trevor Darrell, and Anna
  Rohrbach.
\newblock Women also snowboard: Overcoming bias in captioning models.
\newblock In {\em ECCV}, 2018.

\bibitem{chen2020simple}
Ting Chen, Simon Kornblith, Mohammad Norouzi, and Geoffrey Hinton.
\newblock A simple framework for contrastive learning of visual
  representations.
\newblock In {\em International Conference on Machine Learning}, pages
  1597--1607. PMLR, 2020.

\bibitem{chen2020exploring}
Xinlei Chen and Kaiming He.
\newblock Exploring simple siamese representation learning.
\newblock {\em arXiv preprint arXiv:2011.10566}, 2020.

\bibitem{dosovitskiy2014discriminative}
Alexey Dosovitskiy, Jost~Tobias Springenberg, Martin Riedmiller, and Thomas
  Brox.
\newblock Discriminative unsupervised feature learning with convolutional
  neural networks.
\newblock In {\em Advances in Neural Information Processing Systems}, pages
  766--774, 2014.

\bibitem{dwork2018decoupled}
Cynthia Dwork, Nicole Immorlica, Adam~Tauman Kalai, and Max Leiserson.
\newblock Decoupled classifiers for group-fair and efficient machine learning.
\newblock In {\em Proceedings of the 1st Conference on Fairness, Accountability
  and Transparency}, 2018.

\bibitem{edwards2016censoring}
Harrison Edwards and Amos Storkey.
\newblock Censoring {Representations} with an {Adversary}.
\newblock In {\em ICLR}, 2016.

\bibitem{ganin2015unsupervised}
Yaroslav Ganin and Victor Lempitsky.
\newblock Unsupervised {Domain} {Adaptation} by {Backpropagation}.
\newblock In {\em ICML}, 2015.

\bibitem{gidaris2018unsupervised}
Spyros Gidaris, Praveer Singh, and Nikos Komodakis.
\newblock Unsupervised representation learning by predicting image rotations.
\newblock In {\em International Conference on Learning Representations}, 2018.

\bibitem{grover2019fair}
Aditya Grover, Kristy Choi, Rui Shu, and Stefano Ermon.
\newblock Fair generative modeling via weak supervision.
\newblock {\em arXiv preprint arXiv:1910.12008}, 2019.

\bibitem{grover2019bias}
Aditya {Grover}, Jiaming {Song}, Ashish {Kapoor}, Kenneth {Tran}, Alekh
  {Agarwal}, Eric~J {Horvitz}, and Stefano {Ermon}.
\newblock Bias correction of learned generative models using likelihood-free
  importance weighting.
\newblock In {\em NeurIPS}, 2019.

\bibitem{he_deep_2015}
Kaiming He, Xiangyu Zhang, Shaoqing Ren, and Jian Sun.
\newblock Deep {Residual} {Learning} for {Image} {Recognition}.
\newblock In {\em CVPR}, 2016.

\bibitem{hinton2015distill}
Geoffrey Hinton, Oriol Vinyals, and Jeffrey Dean.
\newblock Distilling the knowledge in a neural network.
\newblock In {\em NIPS Deep Learning and Representation Learning Workshop},
  2014.

\bibitem{khosla2012undoing}
Aditya Khosla, Tinghui Zhou, Tomasz Malisiewicz, Alexei~A Efros, and Antonio
  Torralba.
\newblock Undoing the {Damage} of {Dataset} {Bias}.
\newblock In {\em ECCV}, 2012.

\bibitem{kim2019learning}
Byungju Kim, Hyunwoo Kim, Kyungsu Kim, Sungjin Kim, and Junmo Kim.
\newblock Learning not to learn: Training deep neural networks with biased
  data.
\newblock In {\em CVPR}, 2019.

\bibitem{adam}
Diederik~P. Kingma and Jimmy Ba.
\newblock Adam: {A} method for stochastic optimization.
\newblock {\em CoRR}, abs/1412.6980, 2014.

\bibitem{krizhevsky2009learning}
Alex Krizhevsky, Geoffrey Hinton, et~al.
\newblock Learning multiple layers of features from tiny images.
\newblock 2009.

\bibitem{li2019repair}
Yi Li and Nuno Vasconcelos.
\newblock Repair: Removing representation bias by dataset resampling.
\newblock In {\em CVPR}, 2019.

\bibitem{NEURIPS2020_2288f691}
Hossein Mobahi, Mehrdad Farajtabar, and Peter Bartlett.
\newblock Self-distillation amplifies regularization in hilbert space.
\newblock In H. Larochelle, M. Ranzato, R. Hadsell, M.~F. Balcan, and H. Lin,
  editors, {\em Advances in Neural Information Processing Systems}, volume~33,
  pages 3351--3361. Curran Associates, Inc., 2020.

\bibitem{mobahi2020selfdistillation}
Hossein Mobahi, Mehrdad Farajtabar, and Peter~L. Bartlett.
\newblock Self-distillation amplifies regularization in hilbert space.
\newblock In {\em arXiv preprint arXiv:2002.05715}, 2020.

\bibitem{quadrianto2019discovering}
Novi Quadrianto, Viktoriia Sharmanska, and Oliver Thomas.
\newblock Discovering fair representations in the data domain.
\newblock In {\em CVPR}, 2019.

\bibitem{russakovsky2015imagenet}
Olga Russakovsky, Jia Deng, Hao Su, Jonathan Krause, Sanjeev Satheesh, Sean Ma,
  Zhiheng Huang, Andrej Karpathy, Aditya Khosla, Michael Bernstein, et~al.
\newblock Imagenet large scale visual recognition challenge.
\newblock {\em International journal of computer vision}, 115(3):211--252,
  2015.

\bibitem{ryu2018inclusivefacenet}
Hee~Jung Ryu, Hartwig Adam, and Margaret Mitchell.
\newblock Inclusivefacenet: Improving face attribute detection with race and
  gender diversity.
\newblock In {\em Workshop on Fairness, Accountability, and Transparency in
  Machine Learning (FAT/ML)}, 2018.

\bibitem{ryu2017improving}
Hee~Jung Ryu, Margaret Mitchell, and Hartwig Adam.
\newblock Improving {Smiling} {Detection} with {Race} and {Gender} {Diversity}.
\newblock {\em arXiv preprint arXiv:1712.00193}, 2017.

\bibitem{tzeng2015simultaneous}
Eric Tzeng, Judy Hoffman, Trevor Darrell, and Kate Saenko.
\newblock Simultaneous deep transfer across domains and tasks.
\newblock In {\em CVPR}, 2015.

\bibitem{wang2019racial}
Mei Wang, Weihong Deng, Jiani Hu, Xunqiang Tao, and Yaohai Huang.
\newblock Racial faces in the wild: Reducing racial bias by information
  maximization adaptation network.
\newblock In {\em CVPR}, 2019.

\bibitem{wang2019balanced}
Tianlu Wang, Jieyu Zhao, Mark Yatskar, Kai-Wei Chang, and Vicente Ordonez.
\newblock Balanced datasets are not enough: Estimating and mitigating gender
  bias in deep image representations.
\newblock In {\em CVPR}, 2019.

\bibitem{wang2020towards}
Zeyu Wang, Klint Qinami, Ioannis~Christos Karakozis, Kyle Genova, Prem Nair,
  Kenji Hata, and Olga Russakovsky.
\newblock Towards fairness in visual recognition: Effective strategies for bias
  mitigation.
\newblock In {\em Proceedings of the IEEE/CVF Conference on Computer Vision and
  Pattern Recognition}, pages 8919--8928, 2020.

\bibitem{zemel2013learning}
Rich Zemel, Yu Wu, Kevin Swersky, Toni Pitassi, and Cynthia Dwork.
\newblock Learning {Fair} {Representations}.
\newblock In {\em ICML}, 2013.

\bibitem{zhang2018mitigating}
Brian~Hu Zhang, Blake Lemoine, and Margaret Mitchell.
\newblock Mitigating unwanted biases with adversarial learning.
\newblock In {\em Proceedings of the 2018 AAAI/ACM Conference on AI, Ethics,
  and Society}, 2018.

\bibitem{zhang2016colorful}
Richard Zhang, Phillip Isola, and Alexei~A Efros.
\newblock Colorful image colorization.
\newblock In {\em European Conference on Computer Vision}, pages 649--666.
  Springer, 2016.

\bibitem{zhao_men_2017}
Jieyu Zhao, Tianlu Wang, Mark Yatskar, Vicente Ordonez, and Kai{-}Wei Chang.
\newblock Men {Also} {Like} {Shopping}: {Reducing} {Gender} {Bias}
  {Amplification} using {Corpus}-level {Constraints}.
\newblock In {\em EMNLP}, 2017.

\end{thebibliography}
}

\section{Appendix}
\subsection{Effect of Biased Data on Model Predictions}\label{sec:fullstory} 

\subsubsection{Terminologies used}
\begin{enumerate}
    \item grayscale CIFAR-10S - All images in CIFAR-10S converted to grayscale.
    \item color CIFAR-10S - All images in CIFAR-10S replaced by the corresponding original color images in CIFAR-10.
    \item grayscale L-CIFAR-10S - All images in L-CIFAR-10S converted to grayscale.
    \item color L-CIFAR-10S - All images in L-CIFAR-10S replaced by the corresponding original color images in CIFAR-10.
    \item GrayTrainColorTest (GTCT) on CIFAR-10S - Model trained on grayscale CIFAR-10S and evaluated on a test set containing only color images.
    \item GrayTrainColorTest (GTCT) on L-CIFAR-10S - Model trained on grayscale L-CIFAR-10S and evaluated on a test set containing only color images.
    \item ColorTrainColorTest (CTCT) on CIFAR-10S - Model trained on color CIFAR-10S and evaluated on a test set containing only color images.
    \item ColorTrainColorTest (CTCT) on L-CIFAR-10S - Model trained on color L-CIFAR-10S and evaluated on a test set containing only color images.
\end{enumerate}

\subsubsection{Effect of Data Bias with Sufficient Data}\label{sec:sufdata}

The authors in \cite{wang2020towards} experimentally show that a baseline ResNet-18~\cite{he_deep_2015} model trained on CIFAR-10S achieves $89.0\pm 0.5\%$ classification accuracy on the color test set. Whereas the same model trained on grayscale CIFAR-10S (GTCT) achieves $93.0\%$ classification accuracy on the color test set \cite{wang2020towards}, which is a significant increase. The authors attribute this problem to the skewed data in the CIFAR-10S dataset. As a result, the model is biased towards predicting one among the five color domain classes in the case of colored images since these classes contain mostly color images. The authors in \cite{wang2020towards} propose that the goal of any bias mitigating approach should be to achieve a classification accuracy at least close to that of GTCT on CIFAR-10S ($93.0\%$). It should also ideally approach the classification accuracy of CTCT on CIFAR-10S which is $95.0\%$. The performance gap in the baseline (grayscale CIFAR-10S vs. CIFAR-10S) due to the bias is 4\%. Our approach trained on sufficient data (CIFAR-10S) achieves a classification accuracy of $91.1\%$ on the color test, which is significantly higher than the baseline and reduces this gap from 4\% to around 2\%. This indicates that our approach is effective in mitigating bias even when sufficient training data is present.

The authors in \cite{wang2020towards} compare various bias mitigation approaches on the CIFAR-10S dataset. The experimental results indicate that the domain independent strategy proposed in \cite{wang2020towards} achieves the highest accuracy of 92.4\% on the color test set as compared to the other approaches. The domain discriminative training approach achieves a classification accuracy of 91.2\% on the color test set. These performances are close to the 93\% classification accuracy achieved by GTCT on CIFAR-10S. Therefore, the gap in the performance of the domain discriminative and domain independent training approaches compared to GTCT on the CIFAR-10S dataset are 1.8\% and 0.6\%, respectively. Our approach augmented with the domain independent method trained on CIFAR-10S achieves a classification accuracy of $93.2\%$ on the color test, which outperforms the domain independent (92.4\%) model and also slightly outperforms GTCT (93\%) on CIFAR-10S.

\subsection{Experimental Details}
\subsubsection{Details of the Datasets}
Introductory details regarding the datasets have been provided in Sec.~\ref{sec:dataset} in the main paper. All the CIFAR-10S variants retain the 10 classes and 50,000 training images of the CIFAR-10~\cite{krizhevsky2009learning} dataset but modify the training images to introduce a bias using a domain attribute containing two domains, d1 and d2. The d1 domain has five classes, each containing 95\% d1 type images and 5\% d2 type images (95-5\% skew). The remaining five classes belong to the d2 domain and contain 95\% d2 type and 5\% d1 type images each (95-5\% skew). Further details are given below.

\begin{enumerate}
    \item CIFAR-10S dataset: The domain attribute in this dataset is whether the image is colored (standard CIFAR-10 color images) or grayscale (grayscale version of CIFAR-10 images). The color domain has five classes, each containing 95\% color images and 5\% grayscale images (95-5\% skew). The remaining five classes belong to the grayscale domain and contain 95\% grayscale and 5\% color images each (95-5\% skew). 
    \item CIFAR-10S-i:  The domain attribute in this dataset is whether the image is a standard CIFAR-10 color image or a downsampled image ($32\times32$) of the same class from the ImageNet dataset. The CIFAR-10 color domain has five classes, each containing 95\% color images and 5\% ImageNet images (95-5\% skew). The remaining five classes belong to the ImageNet domain and contain 95\% ImageNet images and 5\% CIFAR-10 color images each (95-5\% skew).
    \item CIFAR-10S-c28: The domain attribute in this dataset is whether the image is a standard CIFAR-10 color image of size $32\times32$ or a CIFAR-10 image cropped to $28\times28$ from the center and upsampled to $32\times32$ (c28). The CIFAR-10 color domain has five classes, each containing 95\% color images and 5\% c28 images (95-5\% skew). The remaining five classes belong to the c28 domain and contain 95\% c28 images and 5\% CIFAR-10 color images each (95-5\% skew).

    \item CIFAR-10S-d16: The domain attribute in this dataset is whether the image is a standard CIFAR-10 color image of size $32\times32$ or a CIFAR-10 image downsampled to $16\times16$ and upsampled to $32\times32$ (d16). The CIFAR-10 color domain has five classes, each containing 95\% color images and 5\% d16 images (95-5\% skew). The remaining five classes belong to the d16 domain and contain 95\% d16 images and 5\% CIFAR-10 color images each (95-5\% skew).

    \item CIFAR-10S-d8: The domain attribute in this dataset is whether the image is a standard CIFAR-10 color image of size $32\times32$ or a CIFAR-10 image downsampled to $8\times8$ and upsampled to $32\times32$ (d8). The CIFAR-10 color domain has five classes, each containing 95\% color images and 5\% d8 images (95-5\% skew). The remaining five classes belong to the d8 domain and contain 95\% d8 images and 5\% CIFAR-10 color images each (95-5\% skew).
   
\end{enumerate}

We have provided concise details regarding the CIFAR-10S, CIFAR-10S-i, CIFAR-10S-c28, CIFAR-10S-d16, and CIFAR-10S-d8 datasets. For further details regarding these datasets, please refer to \cite{wang2020towards}.

The L-CIFAR-10S, L-CIFAR-10S-i, L-CIFAR-10S-c28, L-CIFAR-10S-d16, and L-CIFAR-10S-d8 datasets are limited data versions of CIFAR-10S, CIFAR-10S-i, CIFAR-10S-c28, CIFAR-10S-d16, and CIFAR-10S-d8 datasets, respectively. They contain 5\% of the images of their respective parent dataset but with the same skew level (95-5\%). Specifically, we choose the first 5\% images from each class while maintaining the same level of skew, i.e., we separately choose the first 5\% images from each domain for every class. For example, in the case of L-CIFAR-10S, we separately choose 5\% images from the color and grayscale images in each class. As a result, if a class contained 95\% color images and 5\% grayscale images or vice-versa, the ratio of color and grayscale images will still remain the same in every class of L-CIFAR-10S.

\subsubsection{Details of Compared Approaches}
Introductory details regarding the compared approaches have been provided in Sec.~\ref{sec:compappr} in the main paper. We provide further details regarding the compared approaches below:

\begin{enumerate}
    \item Strategic sampling: In this approach, the rare (minority) images are strategically re-sampled to artificially balance the dataset in terms of the number of images of the two types/domains in each class. However, it also increases the chances of overfitting due to seeing the same images multiple times. It also increases the training time without providing any additional information \cite{wang2020towards}.
    
    \item Adversarial training approach: In this approach, a minimax objective is set to minimize the possibility that the protected attribute can be predicted using the features from the network while maximizing the classification power of the network. It is based on the idea that if the model cannot encode the information regarding the protected attribute, it will not be affected by the bias due to that attribute. In our paper, we perform experiments for both the techniques used in \cite{wang2020towards} under the adversarial approach. Specifically, we perform adversarial training using the uniform confusion loss $-(1/|D|) \sum_{d} \log q_d $ approach, and the loss reversal $ \sum_{d} \mathds{1}[\widehat{d} = d] \log q_d $ with gradient projection approach used in \cite{wang2020towards}.
    
    \item Domain discriminative training approach: In this approach, the protected attribute is explicitly modeled as opposed to the adversarial approach, and the correlation between the classes and the protected attribute is then explicitly removed during inference. The authors in \cite{wang2020towards} employ a simple approach of using an ND-way classifier where N is the number of image classes, and D is the number of domains. They propose three types of inference for this approach. First approach, involves directly adding the probabilities for all the domains per class ($\argmax_y \sum_d \ptr(y,d|x)$). But this does not take into account the prior information regarding the correlation between the classes and domains. The next two approaches involve first applying a prior shift based on this correlation ($\pte(y,d|x) = \ptr(y,d|x)/\ptr(y,d)$) and then either adding the probabilities for all the domains per class ($\argmax_y \sum_d \pte(y,d|x)$) or taking the highest probability without adding the domain wise class probabilities ($\argmax_{y} \max_d \pte(y,d|x)$). The authors in \cite{wang2020towards} also use Reducing Bias Amplification (RBA) \cite{zhao_men_2017} as an inference method. 
    
    \item Domain independent training approach: In this approach, the authors in \cite{wang2020towards} try to avoid the problems in the domain discriminative approach. One such problem is that the domain discriminative approach leads to learning decision boundaries among different domains in the same class, which may be unnecessary, especially in cases when the class prediction is already good. Therefore, in the domain independent training approach, separate classifiers are trained per domain but with a shared feature extraction network. The authors in \cite{wang2020towards} experiment with two inference methods: a) $\hat{y} = \argmax_y \pte(y|\knownd,x)$, if the domain $\knownd$ of the test image is known, b) $\hat{y} = \argmax_y \sum_d \score(y,d,x)$, which is basically the sum of the classification layer activations for each domain per class.
\end{enumerate}

We have provided concise details regarding the bias mitigation methods. For a more detailed discussion regarding these methods, refer to \cite{wang2020towards}.

\subsubsection{Implementation Details} 
For the experiments involving the L-CIFAR-10S dataset and its variants, we train the model from scratch for 200 epochs using an SGD optimizer with an initial learning rate of $1\mathrm{e}-1$, weight decay of $5 \mathrm{e}{-4}$, and momentum of 0.9. The learning rate is decreased by a factor of 10 after every 50 epochs. The training images are padded with 4 pixels, randomly flipped horizontally, and randomly cropped to $32\times32$ \cite{wang2020towards}.  

In the ResNet-50 backbone used for the L-CelebA experiments, the fully connected layer is replaced with 2 fully connected layers \cite{wang2020towards} with a dropout and a ReLU activation layer between them. We train the network using the binary cross-entropy loss for 50 epochs with a batch size of 32. We use the Adam optimizer~\cite{adam} with a learning rate of $1 \mathrm{e}{-4}$.

For the L-CelebA experiments, we use a weighted mAP proposed in \cite{wang2020towards} to remove the gender bias in the test set. If an attribute is more prevalent among the women images, $BA = P_w/(P_m+P_w)-N_w/(N_m+N_w)$ where $P_w,P_m$ refer to the number of images of women and men predicted to have this attribute. $N_w,N_m$ refer to the actual number of women and men images in the training data. If an attribute is more prevalent among the men, $BA = P_m/(P_m+P_w)-N_m/(N_m+N_w)$. Since the objective of bias mitigation is to reduce the level of bias that was learned from the training data and the bias amplification score is expected to be negative as the model becomes fairer across genders \cite{wang2020towards}. 

In our approach, given an image, we apply random augmentations to create two different views. We feed them to the backbone network and perform the standard cross-entropy loss based training for image classification (or binary cross-entropy loss based training for multi-label classification). We also feed the features for both the views to the multi-layer perceptron projection head and then the prediction head in order to apply the SimSiam loss function as described in Eqs.~1, 2, 3, 4 in the main paper. In order to apply self-distillation, we minimize KL divergence between the logits/soft predictions of the student and teacher networks. For all the experiments, the domain labels are assumed to be available during training time. 

We use the same experimental settings and metrics for comparing bias mitigation approaches as used in \cite{wang2020towards}. Please refer to \cite{wang2020towards} for further details.

\end{document}